\documentclass[runningheads]{llncs}
\usepackage{amsmath,amssymb,amsfonts}
\usepackage{multirow}
\usepackage{booktabs}
\usepackage{rotating}
\usepackage{makecell}
\usepackage{algorithmic}
\usepackage{textcomp}
\usepackage{subfigure}

\usepackage{overpic}
\usepackage{graphicx}
\usepackage{xcolor}
\usepackage{bbding} 
\usepackage{rotating,soul}
\usepackage{gensymb}

\usepackage{bbm}

\usepackage{amssymb}
\usepackage{bbding}
\usepackage{pifont}
\usepackage{wasysym}
\usepackage{utfsym}

\usepackage{algorithmic}
\usepackage{textcomp}
\usepackage[misc]{ifsym}

\begin{document}
\title{Towards Novel Class Discovery: A Study in Novel Skin Lesions Clustering}
\author{Wei Feng\inst{1,2,4,6} \and 
Lie Ju\inst{1,2,4,6}\and 
Lin Wang\inst{1,2,4,6}\and 
Kaimin Song\inst{6} \and 
Zongyuan Ge\inst{1,2,3,4,5}$^{(\textrm{\Letter})}$} 
\index{Feng, Wei}
\index{Ju, Lie}
\index{Wang, Lin}

\index{Song, Kaimin}
\index{Ge, Zongyuan}
%
\institute{Faculty of Engineering, Monash University, Melbourne, Australia\and 
Monash Medical AI Group, Monash University, Melbourne, Australia\and
AIM for Health Lab, Monash University, Victoria, Australia \and
Airdoc-Monash Research Lab, Monash University, Suzhou, China 
 \and Faculty of IT, Monash University, Victoria, Australia \and
Airdoc LLC, Beijing, China \\
\email{wf02429@gmail.com}, \email{zongyuan.ge@monash.edu}\\
\url{https://www.monash.edu/mmai-group} \\
}
\maketitle            
\begin{abstract}
Existing deep learning models have achieved promising performance in recognizing skin diseases from dermoscopic images.
However, these models can only recognize samples from predefined categories, when they are deployed in the clinic, data from new unknown categories are constantly emerging. 
Therefore, it is crucial to automatically discover and identify new semantic categories from new data. In this paper, we propose a new novel class discovery framework for automatically discovering new semantic classes from dermoscopy image datasets based on the knowledge of known classes. Specifically, we first use contrastive learning to learn a robust and unbiased feature representation based on all data from known and unknown categories. 
We then propose an uncertainty-aware multi-view cross pseudo-supervision strategy, which is trained jointly on all categories of data using pseudo labels generated by a self-labeling strategy.
Finally, we further refine the pseudo label by aggregating neighborhood information through local sample similarity to improve the clustering performance of the model for unknown categories. We conducted extensive experiments on the dermatology dataset ISIC 2019, and the experimental results show that our approach can effectively leverage knowledge from known categories to discover new semantic categories. We also further validated the effectiveness of the different modules through extensive ablation experiments. Our code will be released soon.

\end{abstract}


\section{Introduction}
Automatic identification of lesions from dermoscopic images is of great importance for the diagnosis of skin cancer \cite{zhang2019attention,mahbod2019fusing}. Currently, deep learning models, especially those based on deep convolution neural networks, have achieved remarkable success in this task \cite{zhang2019attention,tang2020gp,yao2021single}.
However, this comes at the cost of a large amount of labeled data that needs to be collected for each class. To alleviate the labeling burden, semi-supervised learning has been proposed to exploit a large amount of unlabeled data to improve performance in the case of limited labeled data \cite{liu2022acpl,you2022simcvd,hang2020local}. However, it still requires a small amount of labeled data for each class, which is often impossible in real practice. 
For example, there are roughly more than 2000 named dermatological diseases today, of which more than 200 are common, and new dermatological diseases are still emerging, making it impractical to annotate data from scratch for each new disease category \cite{zhang2021opportunities}.
However, since there is a correlation between new and known diseases, a priori knowledge from known diseases is expected to help automatically identify new diseases \cite{han2021autonovel}.

One approach to address the above problem is novel class discovery (NCD) \cite{han2021autonovel,zhong2021openmix,fini2021unified}, which aims to transfer knowledge from known classes to discover new semantic classes.
Most NCD methods follow a two-stage scheme: 1) a stage of fully supervised training on known category data and 2) a stage of clustering on unknown categories \cite{han2021autonovel,zhong2021openmix,fini2021unified}. For example, Han et al. \cite{han2021autonovel} further introduced self-supervised learning in the first stage to learn general feature representations. They also used ranking statistics to compute pairwise similarity for clustering.
Zhong et al. \cite{zhong2021openmix} proposed OpenMix  based on the mixup strategy \cite{zhang2017mixup} to further exploit the information from known classes to improve the performance of unsupervised clustering. Fini et al. \cite{fini2021unified} proposed UNO, which unifies multiple objective functions into a holistic framework to achieve better interaction of information between known and unknown classes. Zhong  et al. \cite{zhong2021neighborhood} used neighborhood information in the embedding space to learn more discriminative representations.
However, most of these methods require the construction of a pairwise similarity prediction task to perform clustering based on pairwise similarity pseudo labels between samples. In this process, the generated pseudo labels are usually noisy, which may affect the clustering process and cause error accumulation. In addition, they only consider the global alignment of samples to the category center, ignoring the local inter-sample alignment thus leading to poor clustering performance.

In this paper, we propose a new novel class discovery framework to automatically discover novel disease categories. Specifically, we first use contrastive learning to pretrain the model based on all data from known and unknown categories to learn a robust and general semantic feature representation. Then, we propose an uncertainty-aware multi-view cross-pseudo-supervision strategy to perform clustering. It first uses a self-labeling strategy to generate pseudo-labels for unknown categories, which can be treated homogeneously with ground truth labels. The cross-pseudo-supervision strategy is then used to force the model to maintain consistent prediction outputs for different views of unlabeled images. In addition, we propose to use prediction uncertainty to adaptively adjust the contribution of the pseudo labels to mitigate the effects of noisy pseudo labels. Finally, to encourage local neighborhood alignment and further refine the pseudo labels, we propose a  local information aggregation module to aggregate the information of the neighborhood samples to boost the clustering performance. We conducted extensive experiments on the dermoscopy dataset ISIC 2019, and the experimental results show that our method outperforms other state-of-the-art comparison algorithms by a large margin. In addition, we also validated the effectiveness of different components through extensive ablation experiments.
\section{METHODOLOGY}

Given an unlabeled dataset $\{x_i^u\}_{i=1}^{N^u}$ with $N^u$ images, where $x_i^u$ is the $i$th unlabeled image. Our goal is to automatically cluster the unlabeled data into $C^u$ clusters. In addition, we also have access to a labeled dataset $\{x_i^l, y_i^l\}_{i=1}^{N^l}$  with $N^l$ images, where $x_i^l$ is the $i$th labeled image and $y_i^l \in \mathcal{Y}=\left\{1, \ldots, C^l\right\}$ is its corresponding label. In the novel class discovery task, the known and unknown classes are disjoint, i.e., $C^{l} \cap C^{u}=\varnothing$. However, the known and unknown classes are similar, and we aim to use the knowledge of the known classes to help the clustering of the unknown classes.
\begin{figure*}[t!]
    \centering
        \includegraphics[width=0.9\linewidth]{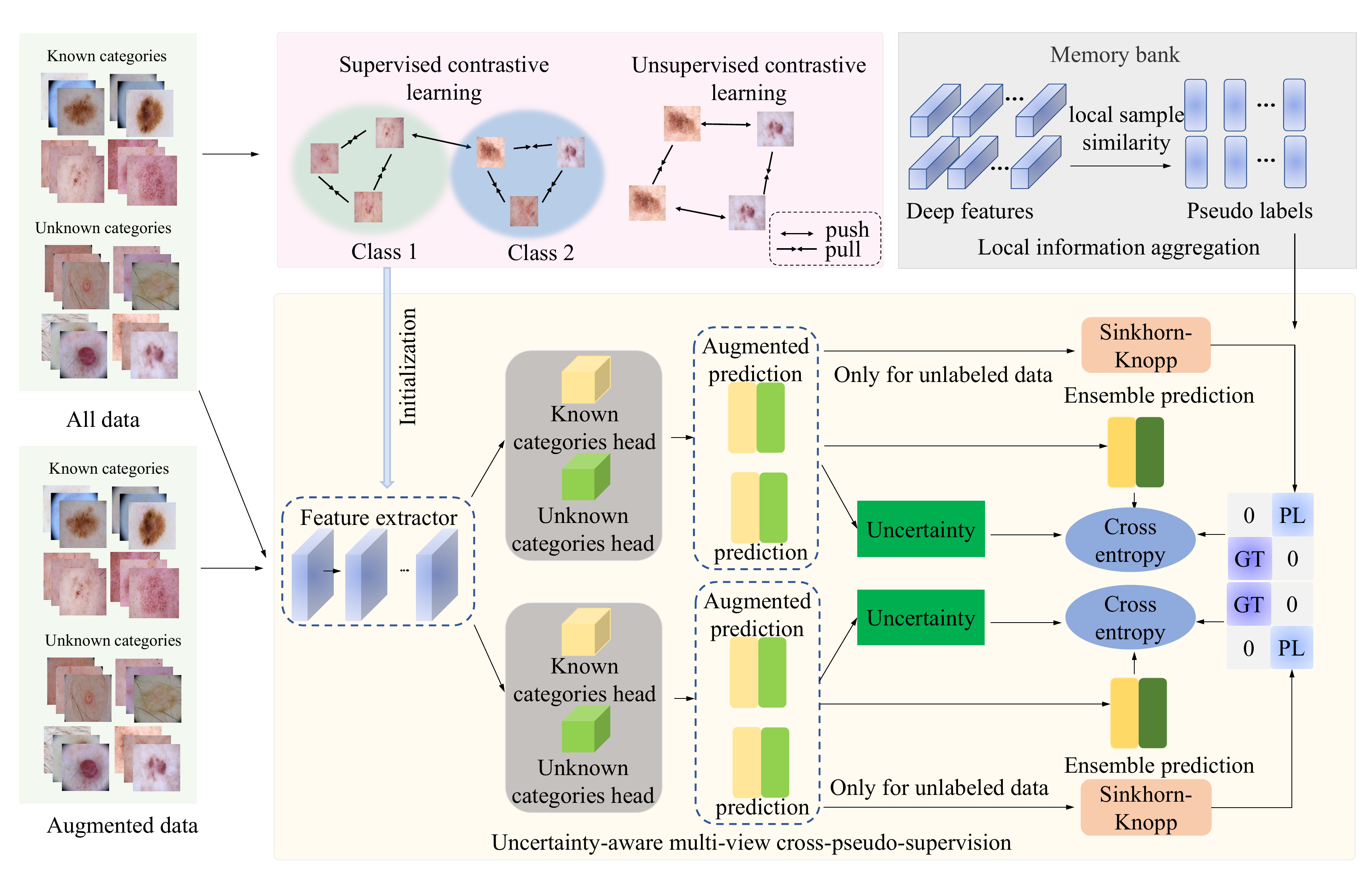} 

    \caption{The overall framework of our proposed novel class discovery   algorithm.}
    \label{fig:framework}
\end{figure*}
The overall framework of our proposed novel class discovery  algorithm is shown in Fig.~\ref{fig:framework}. Specifically, we first learn general and robust feature representations through contrastive learning. Then, the uncertainty-aware multi-view cross-pseudo-supervision strategy is used for joint training on all category data. Finally, the local information aggregation module benefits the NCD by aggregating the useful information of the neighborhood samples.

\subsubsection{Contrastive Learning}
To achieve a robust feature representation for the NCD task, we first use noise contrastive learning \cite{gutmann2010noise} to pretrain the feature extractor network, which effectively avoids model over-fitting to known categories. Specifically, we use $x_{i}$ and $x_{i}^{\prime}$ to represent different augmented versions of the same image in a mini-batch. The unsupervised contrastive loss can be formulated as:
\begin{equation}\label{equ:ucl}
L_{i}^{ucl}=-\log \frac{\exp \left(z_{i} \cdot z_{i}^{\prime} / \tau\right)}{\sum_{n} \mathbbm{1}_{[n \neq i]} \exp \left(z_{i} \cdot z_{n} / \tau\right)}
\end{equation}
where $z_{i}=E(x_{i})$ is the deep feature representation of the image $x_{i}$, $E$ is the feature extractor network, and $\tau$ is the temperature value. $\mathbbm{1}$ is the indicator function.

In addition, to help the feature extractor learn semantically meaningful feature representations, we introduce supervised contrastive learning \cite{khosla2020supervised} for labeled known category data, which can be denoted as:
\begin{equation}\label{equ:scl}
L_{i}^{scl}=-\frac{1}{|N(i)|} \sum_{q \in N(i)} \log \frac{\exp \left(z_{i} \cdot z_{q} / \tau\right)}{\sum_{n} \mathbbm{1}_{[n \neq i]} \exp \left(z_{i} \cdot z_{n} / \tau\right)}
\end{equation}
where $N(i)$ represents the sample set with the same label as $x_{i}$ in a mini-batch data. $|N(i)|$ represents the number of samples.

The overall contrastive loss can be expressed as:
$L_{cl}= (1-\mu) \sum_{i \in B} L_{i}^{ucl}+\mu \sum_{i \in B_{l}} L_{i}^{scl}$
,where $\mu$ denotes the balance coefficient. $B_{l}$ is the labeled subset of mini-batch data. 
\subsubsection{Uncertainty-aware multi-view cross-pseudo-supervision}
We now describe how to train uniformly on known and unknown categories using the uncertainty-aware multi-view cross-pseudo-supervision strategy.  Specifically, we construct two parallel classification models $M_1$ and $M_2$, both of them composed of a feature extractor and two category classification heads, using different initialization parameters. For an original image $x_i$, we generate two augmented versions of $x_i$, $x_i^{v1}$ and $x_i^{v2}$. We then feed these two augmented images into $M_1$ and $M_2$ to obtain the predictions for $x_i^{v1}$ and $x_i^{v2}$:
\begin{equation}\label{equ:vp}
\begin{array}{l}
p^{v1}_{i,1}=M_1(x_i^{v1}), 
p^{v2}_{i,1}=M_1(x_i^{v2}), 
p^{v1}_{i,2}=M_2(x_i^{v1}), 
p^{v2}_{i,2}=M_2(x_i^{v2}) .
\end{array}
\end{equation}

The prediction outputs are obtained by concatenating the outputs of the two classification heads and then passing a softmax layer \cite{fini2021unified}. Then, we can compute the ensemble predicted output of $M_1$ and $M_2$:
$p_{i}^{M_1}=\left(p^{v1}_{i,1}+p^{v2}_{i,1}\right) / 2$,
$p_{i}^{M_2}=\left(p^{v1}_{i,2}+p^{v2}_{i,2}\right) / 2$.

Next, we need to obtain training targets for all data. For an input image $x_i$, if $x_i$ is from the known category, we construct the training target as one hot vector, where the first $C^l$ elements are ground truth labels and the last $C^u$ elements are 0. If $x_i$ is from the unknown category, we set the first $C^l$ elements to 0 and use pseudo labels for the remaining $C^u$ elements.

We follow the self-labeling method in \cite{asano2019self,caron2018deep} to generate pseudo labels.
Specifically, the parameters in the unknown category classification head can be viewed as prototypes of each category, and our training goal is to distribute a set of samples uniformly to each prototype while maximizing the similarity between samples and prototypes \cite{asano2019self}. Let $\mathbf{P}=\left[p_{1}^{u} ; \ldots ; p_{B_{u}}^{u}\right] \in \mathbb{R}^{B_{u} \times C^u}$ denotes the ensemble prediction of data of unknown categories in a mini-batch, where $B_{u}$ represents the number of samples. Here we only consider the output of the unknown categories head due to the samples coming from unknown categories \cite{fini2021unified}. We obtain the pseudo label by optimizing the following objective:
\begin{equation}\label{equ:gp}
\max _{\mathbf{Y} \in \mathcal{S}} \operatorname{tr}\left(\mathbf{Y} \mathbf{P}^{\top} \right)+\delta H(\mathbf{Y})
\end{equation}
where $\mathbf{Y}=\left[y_{1}^{u} ; \ldots ; y_{B_{u}}^{u}\right] \in \mathbb{R}^{B_{u} \times C^{u}}$ will assign $B_{u}$ unknown category samples to $C^{u}$ category prototypes uniformly, i.e., each category prototype will be selected $B_{u}/C^{u}$ times on average. $\mathcal{S}$ is the search space. $H$ is the entropy function used to control the smoothness of $\mathbf{Y}$. $\delta$ is the hyperparameter. The solution to this objective can be calculated by the Sinkhorn-Knopp algorithm \cite{cuturi2013sinkhorn}.
After generating the pseudo-labels, we can combine them with the ground truth labels of known categories as training targets for uniform training.

To mitigate the effect of noisy pseudo labels, we propose to use prediction uncertainty \cite{li2022learning} to adaptively adjust the weights of pseudo labels. Specifically, we first compute the variance of the predicted outputs of the models for the different augmented images via KL-divergence:
\begin{equation}\label{equ:va}
\begin{array}{l}
V_{1}=E\left[p^{v1}_{i,1} \log \left(\frac{p^{v1}_{i,1}}{p^{v2}_{i,1}}\right)\right], 
V_{2}=E\left[p^{v1}_{i,2} \log \left(\frac{p^{v1}_{i,2}}{p^{v2}_{i,2}}\right)\right],
\end{array}
\end{equation}
where $E$ represents the expected value. If the variance of the model's predictions for different augmented images is large, the pseudo label may be of low quality, and vice versa. Then, based on the prediction variance of the two models, the multi-view cross-pseudo supervision loss can be formulated as:
\begin{equation}\label{equ:cpsloss}
L_{cps}=E\left[e^{-V_{1}} L_{c e}\left(p^{M_2}, y^{v1}\right)+V_{1}\right]+E\left[e^{-V_{2}} L_{c e}\left(p^{M_1}, y^{v2}\right)+V_{2}\right]
\end{equation}
where $L_{c e}$ denotes the cross-entropy loss. $y^{v1}$ and $y^{v2}$ are the training targets.
\subsubsection{Local information aggregation}

After the cross-pseudo-supervision training described above, we are able to assign the instances to their corresponding clustering centers. However, it ignores the alignment between local neighborhood samples, i.e., the samples are susceptible to interference from some irrelevant semantic factors such as background and color. Here, we propose a local information aggregation to enhance the alignment of local samples. Specifically, as shown in Fig.~\ref{fig:framework}, we maintain a first-in-first-out memory bank $\mathcal{M}=\left\{z_{k}^{m}, y_{k}^{m}\right\}_{k=1}^{N^{m}}$ during the training process, which contains the features of $N^m$ most recent samples and their pseudo labels. For each sample in the current batch, we compute the similarity between its features and the features of each sample in the memory bank:
\begin{equation}\label{equ:local}
d_{k}=\frac{\exp \left(z \cdot z_{k}^{m} \right)}{\sum_{k=1}^{N^{m}} \exp \left(z \cdot z_{k}^{m}  \right)}.
\end{equation}

Then based on this feature similarity, we obtain the final pseudo labels as:
$y^{u}=\rho y^{u}+(1-\rho) \sum_{k=1}^{N^{m}} d_{k} y_{k}^{m}$,
where $\rho$ is the balance coefficient. By aggregating the information of the neighborhood samples, we are able to ensure consistency between local samples, which further improves the clustering performance.


\section{Experiments}
\subsubsection{Dataset.}
To validate the effectiveness of the proposed algorithm, we conduct experiments on the widely used public dermoscopy challenge dataset ISIC 2019 \cite{codella2018skin,combalia2019bcn20000}. The dataset contains a total of 25,331 dermoscopic images from eight categories: Melanoma (MEL), Melanocytic Nevus (NV), Basal Cell Carcinoma (BCC), Actinic Keratosis
(AK), Benign Keratosis (BKL), Dermatofibroma (DF),
Vascular Lesion (VASC), and Squamous Cell Carcinoma
(SCC). Since the dataset suffers from severe category imbalance, we randomly sampled 500 samples from those major categories (MEL, NV, BCC, BKL) to maintain category balance. Then, we construct the NCD task where we treat 50\% of the categories (AK, MEL, NV, BCC) as known categories and the remaining 50\% of the categories (BKL, SCC, DF, VASC) as unknown categories. We also swap the known and unknown categories to form a second NCD task. For task 1 and task 2, we report the average performance of 5 runs.

\subsubsection{Implementation details.}
We used ResNet-18 \cite{he2016deep} as the backbone of the classification model. The known category classification head is an \textit{l2}-normalized linear classifier with $C^l$ output units. The unknown category classification head consists of a projection layer with 128 output units, followed by an \textit{l2}-normalized linear classifier with $C^u$ output units. In the first contrastive learning pre-training step, we used SGD optimizer to train the model for 200 epochs and gradually decay the learning rate starting from 0.1 and dividing it by 5 at the epochs 60, 120, and 180. $\mu$ is set to 0.5, $\tau$ is set to 0.5.  In the joint training phase, we fix the parameters of the previous feature extractor and only fine-tune the parameters of the classification head. We use the SGD optimizer to train the model for 200 epochs with linear warm-up and cosine annealing ($l r_{\text {base}}=0.1$, $l r_{\text {min}}=0.001$), and the weight decay is set to $1.5 \times 10^{-4}$.  For data augmentation, we use random horizontal/vertical flipping, color jitter, and Gaussian blurring following \cite{fini2021unified}. For pseudo label, we use the Sinkhorn-Knopp algorithm with hyperparameters inherited from \cite{fini2021unified}: $\delta=0.05$ and the number of iterations is 3. We use a memory bank $\mathcal{M}$ of size 100 and the hyperparameter $\rho$ is set to 0.6. The batch size in all experiments is 512. In the inference phase, we only use the output of the unknown category classification head of $M_1$ \cite{han2021autonovel}. Following \cite{han2021autonovel,zhong2021neighborhood,zhong2021openmix}, we report the clustering performance on the unlabeled unknown category dataset. We assume that the number of unknown categories is known and it can also be obtained by the category number estimation method proposed in \cite{han2021autonovel}.

Following \cite{cao2021open,han2021autonovel}, we use the average clustering accuracy (ACC), normalized mutual information (NMI) and adjusted rand index (ARI) to evaluate the clustering performance of different algorithms. Specifically, we first match the clustering assignment and ground truth labels by the Hungarian algorithm \cite{kuhn1955hungarian}. After the optimal assignment is determined, we then compute each metric. 
We implement all algorithms based on the PyTorch framework and conduct experiments on 8 RTX 3090 GPUs.
\subsubsection{Comparison with state-of-the-art methods.}
We compare our algorithms with some state-of-the-art NCD methods, including RankStats \cite{han2021autonovel}, RankStats+ (RankStats with incremental learning) \cite{han2021autonovel}, OpenMix \cite{zhong2021openmix}, NCL \cite{zhong2021neighborhood}, UNO \cite{fini2021unified}. we also compare with the benchmark method (Baseline), which first trains a model using known category data and then performs clustering on unknown category data. Tabel~\ref{tab:compare} shows the clustering performance of each comparison algorithm on different NCD tasks. It can be seen that the clustering performance of the benchmark method is poor, which indicates that the model pre-trained using only the known category data does not provide a good clustering of the unknown category. Moreover, the state-of-the-art NCD methods can improve the clustering performance, which demonstrates the effectiveness of the currently popular two-stage solution. However, our method outperforms them, mainly due to the fact that they need to generate pairwise similarity pseudo labels through features obtained based on self-supervised learning, while ignoring the effect of noisy pseudo labels. Compared with the best comparison algorithm UNO, our method yields 5.23\% ACC improvement, 3.56\% NMI improvement, and 2.55\% ARI improvement on Task1, and 3.24\% ACC improvement, 1.34\% NMI improvement, and 2.37\% ARI improvement on Task2, which shows that our method is able to provide more reliable pseudo labels for NCD.

\begin{table*}[t!]
\centering
    \caption{Clustering performance of different comparison algorithms on different tasks.}
     \label{tab:compare}
     \setlength{\tabcolsep}{3mm}
\scriptsize
\begin{tabular}{lcccccc}
\hline
\multirow{2}{*}{Method} & \multicolumn{3}{c}{Task1}                           & \multicolumn{3}{c}{Task2}                           \\  \cmidrule(lr){2-4} \cmidrule(lr){5-7} 
                        & ACC             & NMI             & ARI             & ACC             & NMI             & ARI             \\ \hline
Baseline                & 0.4685          & 0.2107          & 0.1457          & 0.3899          & 0.0851          & 0.0522          \\
RankStats \cite{han2021autonovel}              & 0.5652          & 0.2571          & 0.2203          & 0.4284          & 0.1164          & 0.1023          \\
RankStats+ \cite{han2021autonovel}             & 0.5845          & 0.2633          & 0.2374          & 0.4362          & 0.1382          & 0.1184          \\
OpenMix \cite{zhong2021openmix}                  & 0.6083          & 0.2863          & 0.2512          & 0.4684          & 0.1519          & 0.1488          \\
NCL   \cite{zhong2021neighborhood}                  & 0.5941          & 0.2802          & 0.2475          & 0.4762          & 0.1635          & 0.1573          \\
UNO  \cite{fini2021unified}                    & 0.6131          & 0.3016          & 0.2763          & 0.4947          & 0.1692          & 0.1796          \\
Ours                    & \textbf{0.6654} & \textbf{0.3372} & \textbf{0.3018} & \textbf{0.5271} & \textbf{0.1826} & \textbf{0.2033} \\ \hline
\end{tabular}
\end{table*}

\subsubsection{Ablation study of each key component.}

We performed ablation experiments to verify the effectiveness of each component. As shown in Tabel~\ref{tab:key}, CL is contrastive learning, UMCPS is uncertainty-aware multi-view cross-pseudo-supervision, and LIA is the local information aggregation module. It can be observed that CL brings a significant performance gain, which indicates that contrastive learning helps to learn a general and robust feature representation for NCD. In addition, UMCPS also improves the clustering performance of the model, which indicates that unified training helps to the category information interaction. LIA further improves the clustering performance, which indicates that local information aggregation helps to provide better pseudo labels. Finally, our algorithm incorporates each component to achieve the best performance.
\begin{table*}[t!]
\centering
    \caption{Ablation study of each key component.}
     \label{tab:key}
     \setlength{\tabcolsep}{2mm}
\scriptsize
\begin{tabular}{ccccccccc}
\hline
\multicolumn{3}{c}{Method}                         & \multicolumn{3}{c}{Task1}                           & \multicolumn{3}{c}{Task2}                           \\ \cmidrule(lr){1-3} \cmidrule(lr){4-6} \cmidrule(lr){7-9} 
CL & UMCPS                 & LIA                   & ACC             & NMI             & ARI             & ACC             & NMI             & ARI             \\ \hline
\usym{2717}  & \multicolumn{1}{c}{\usym{2717}} & \multicolumn{1}{c}{\usym{2717}} & 0.4685          & 0.2107          & 0.1457          & 0.3899          & 0.0851          & 0.0522          \\
\usym{2713}  &                       &                       & 0.5898          & 0.2701          & 0.2375          & 0.4402          & 0.1465          & 0.1322          \\
\usym{2713}  & \usym{2713}                    &                       & 0.6471          & 0.3183          & 0.2821          & 0.5012          & 0.1732          & 0.1851          \\
\usym{2713}  &                       & \usym{2713}                    & 0.6255          & 0.3122          & 0.2799          & 0.4893          & 0.1688          & 0.1781          \\
\usym{2713}  & \usym{2713}                   & \usym{2713}                    & \textbf{0.6654} & \textbf{0.3372} & \textbf{0.3018} & \textbf{0.5271} & \textbf{0.1826} & \textbf{0.2033} \\ \hline
\end{tabular}
\end{table*}
\begin{table*}[t!]
\centering
    \caption{Ablation study of contrastive learning and  uncertainty-aware multi-view cross-pseudo-supervision.}
     \label{tab:ab1}
     \setlength{\tabcolsep}{3mm}
\scriptsize
\begin{tabular}{lcccccc}
\hline
\multirow{2}{*}{Method} & \multicolumn{3}{c}{Task1}                           & \multicolumn{3}{c}{Task2}                           \\\cmidrule(lr){2-4} \cmidrule(lr){5-7} 
                        & ACC             & NMI             & ARI             & ACC             & NMI             & ARI             \\ \hline
Baseline                & 0.4685          & 0.2107          & 0.1457          & 0.3899          & 0.0851          & 0.0522          \\
SCL                     & 0.5381          & 0.2362          & 0.1988          & 0.4092          & 0.1121          & 0.1003          \\
UCL                     & 0.5492          & 0.2482          & 0.2151          & 0.4291          & 0.1173          & 0.1174          \\
SCL+UCL                 & \textbf{0.5898} & \textbf{0.2701} & \textbf{0.2375} & \textbf{0.4402} & \textbf{0.1465} & \textbf{0.1322} \\ \hline

w/o CPS                 & 0.6021          & 0.2877          & 0.2688          & 0.4828          & 0.1672          & 0.1629          \\
CPS                     & 0.6426          & 0.3201          & 0.2917          & 0.5082          & 0.1703          & 0.1902          \\
UMCPS                   & \textbf{0.6654} & \textbf{0.3372} & \textbf{0.3018} & \textbf{0.5271} & \textbf{0.1826} & \textbf{0.2033} \\ \hline
\end{tabular}

\end{table*}

\subsubsection{Ablation study of contrastive learning.}
We further examined the effectiveness of each component in contrastive learning. Recall that the contrastive learning strategy includes supervised contrastive learning for the labeled known category data and unsupervised contrastive learning for all data. As shown in Tabel~\ref{tab:ab1}, it can be observed that both components improve the clustering performance of the model, which indicates that SCL helps the model to learn semantically meaningful feature representations, while UCL makes the model learn robust unbiased feature representations and avoid its overfitting to known categories.


\subsubsection{Uncertainty-aware multi-view cross-pseudo-supervision.}
We also examine the effectiveness of uncertainty-aware multi-view cross-pseudo-supervision. We compare it with 1) w/o CPS, which does not use cross-pseudo-supervision, and 2) CPS, which uses cross-pseudo-supervision but not the uncertainty to control the contribution of the pseudo label. As shown in Tabel~\ref{tab:ab1}, it can be seen that CPS outperforms w/o CPS, which indicates that CPS encourages the model to maintain consistent predictions for different augmented versions of the input images, and enhances the generalization performance of the model. UMCPS achieves the best clustering performance, which shows its ability to use uncertainty to alleviate the effect of noisy pseudo labels and avoid causing error accumulation.

\section{Conclusion}
In this paper, we propose a novel class discovery framework for discovering new dermatological classes. Our approach consists of three key designs. First, contrastive learning is used to learn a robust feature representation. Second, uncertainty-aware multi-view cross-pseudo-supervision strategy is trained uniformly on data from all categories, while prediction uncertainty is used to alleviate the effect of noisy pseudo labels. Finally, the local information aggregation module further refines the pseudo label by aggregating the neighborhood information to improve the clustering performance. Extensive experimental results validate the effectiveness of our approach. Future work will be to apply this framework to other medical image analysis tasks.

\bibliographystyle{splncs04}
\bibliography{ref}
\end{document}


\title{Towards Novel Class Discovery: A Study in Novel Skin Lesions Clustering}
\author{Paper ID: 283}
\institute{Anonymous Organization \\
\email{**@******.***}
}
\maketitle            

\begin{table*}[htbp!]
\centering
\caption{Dataset statistics for different tasks after resampling}
\label{tab:my-table}
\setlength{\tabcolsep}{3mm}
\begin{tabular}{lcccccccc}
\hline
\multicolumn{9}{c}{Task1}                                                                 \\ \hline
      & \multicolumn{4}{c|}{Known classes}          & \multicolumn{4}{c}{Unknown classes} \\ \hline
class & AK  & MEL & NV  & \multicolumn{1}{l|}{BCC}  & BKL     & SCC    & DF     & VASC    \\
num   & 867 & 500 & 500 & \multicolumn{1}{l|}{500}  & 500     & 628    & 239    & 253     \\ \hline
\multicolumn{9}{c}{Task2}                                                                \\ \hline
      & \multicolumn{4}{c|}{Known classes}          & \multicolumn{4}{c}{Unknown classes} \\ \hline
class & BKL & SCC & DF  & \multicolumn{1}{l|}{VASC} & AK      & MEL    & NV     & BCC     \\
num   & 500 & 628 & 239 & \multicolumn{1}{l|}{253}  & 867     & 500    & 500    & 500     \\ \hline
\end{tabular}
\end{table*}

\begin{table}[htbp!]
\centering
\caption{Clustering performance of different algorithms on the task agnostic evaluation paradigm. We report the performance of the algorithms on known classes, unknown classes, and all classes.}
\label{tab:my-table2}
\setlength{\tabcolsep}{3mm}
\begin{tabular}{lcccccc}
\hline
\multicolumn{7}{c}{Task-agnostic}                                                                                                      \\ \hline
\multicolumn{1}{c}{Task}   & \multicolumn{3}{c}{Task1}                           & \multicolumn{3}{c}{Task2}                           \\ \hline
\multicolumn{1}{c}{Method} & Known           & Unknown         & All             & Known           & Unknown         & All             \\ \hline
Baseline                   & 0.7873          & 0.4472          & 0.6092          & 0.8163          & 0.3572          & 0.5873          \\
RS                         & 0.8071          & 0.5466          & 0.6551          & 0.8452          & 0.4078          & 0.6299          \\
RS+                        & 0.8154          & 0.5711          & 0.6777          & 0.8546          & 0.4162          & 0.6452          \\
OpenMix                    & 0.8362          & 0.5973          & 0.6983          & 0.8673          & 0.4599          & 0.6673          \\
NCL                        & 0.8261          & 0.5619          & 0.6898          & 0.8782          & 0.4563          & 0.6762          \\
UNO                        & 0.8599          & 0.6082          & 0.7227          & 0.8892          & 0.4783          & 0.6902          \\
Ours                       & \textbf{0.8885} & \textbf{0.6463} & \textbf{0.7679} & \textbf{0.9027} & \textbf{0.5094} & \textbf{0.7101} \\ \hline
\end{tabular}
\end{table}

\begin{table}[htbp!]
\centering
\caption{Parametric sensitivity analysis on the parameter $\rho$.}
\label{tab:my-table3}
\setlength{\tabcolsep}{3mm}
\begin{tabular}{lcccccc}
\hline
Task  & \multicolumn{3}{c}{Task1}                           & \multicolumn{3}{c}{Task2}                           \\ \hline
$\rho$ & ACC             & NMI             & ARI             & ACC             & NMI             & ARI             \\ \hline
0.2   & 0.6528          & 0.3201          & 0.2902          & 0.5123          & 0.1721          & 0.1913          \\
0.4   & 0.6628          & 0.3301          & 0.3001          & 0.5178          & 0.1761          & 0.1921          \\
0.6   & \textbf{0.6654} & \textbf{0.3372} & \textbf{0.3016} & \textbf{0.5271} & \textbf{0.1826} & \textbf{0.2033} \\
0.8   & 0.6583          & 0.3292          & 0.2988          & 0.5208          & 0.1782          & 0.1988          \\ \hline
\end{tabular}
\end{table}

\begin{table}[htbp!]
\centering
\caption{Parametric sensitivity analysis on the size of $\mathcal{M}$.}
\label{tab:my-table4}
\setlength{\tabcolsep}{3mm}
\begin{tabular}{lcccccc}
\hline
Task & \multicolumn{3}{c}{Task1}                           & \multicolumn{3}{c}{Task2}                           \\ \hline
$\mathcal{M}$    & ACC             & NMI             & ARI             & ACC             & NMI             & ARI             \\ \hline
50   & 0.6621          & 0.3294          & 0.3001          & 0.5202          & 0.1811          & 0.2004          \\
100  & 0.6654          & 0.3372          & 0.3016          & 0.5271          & 0.1826          & 0.2033          \\
150  & \textbf{0.6658} & \textbf{0.3379} & \textbf{0.3023} & \textbf{0.5278} & \textbf{0.1829} & \textbf{0.2038} \\
200  & 0.6656          & 0.3376          & 0.3021          & 0.5274          & 0.1827          & 0.2035          \\ \hline
\end{tabular}
\end{table}

\begin{table}[htbp!]
\centering
\caption{Parametric sensitivity analysis on the parameter $\tau$.}
\label{tab:my-table5}
\setlength{\tabcolsep}{3mm}
\begin{tabular}{lcccccc}
\hline
Task & \multicolumn{3}{c}{Task1}                           & \multicolumn{3}{c}{Task2}                           \\ \hline
$\tau$    & ACC             & NMI             & ARI             & ACC             & NMI             & ARI             \\ \hline
0.3  & 0.5719          & 0.2643          & 0.2253          & 0.4341          & 0.1373          & 0.1299          \\
0.5  & \textbf{0.5898} & \textbf{0.2701} & \textbf{0.2375} & \textbf{0.4402} & \textbf{0.1465} & \textbf{0.1322} \\
0.8  & 0.5865          & 0.2657          & 0.2303          & 0.4389          & 0.1402          & 0.1305          \\
1.2  & 0.5801          & 0.2648          & 0.2301          & 0.4393          & 0.1414          & 0.1317          \\ \hline
\end{tabular}
\end{table}

\begin{table}[htbp!]
\centering
\caption{Estimation of the number of classes in unlabelled data.}
\label{tab:my-table6}
\begin{tabular}{lcc}
\hline
             & Task1 & Task2 \\ \hline
Ground truth & 4     & 4     \\
Ours         & 3     & 3     \\ \hline
\end{tabular}
\end{table}